\documentclass{article}

\usepackage{graphicx}%
\usepackage{multirow}%
\usepackage{amsmath,amssymb,amsfonts}%
\usepackage{amsthm}%
\usepackage{mathrsfs}%
\usepackage[title]{appendix}%
\usepackage{xcolor}%
\usepackage{textcomp}%
\usepackage{manyfoot}%
\usepackage{booktabs}%
\usepackage{algorithm}%
\usepackage{algorithmicx}%
\usepackage{algpseudocode}%
\usepackage{listings}%

\usepackage{makecell}
\usepackage{setspace}
\usepackage{url}
\usepackage{authblk}

\title{Identification of Regulatory Requirements Relevant to Business Processes:\\ A Comparative Study on Generative AI, Embedding-based Ranking, Crowd and Expert-driven Methods} 

\author[1]{Catherine Sai}
\author[2]{Shazia Sadiq}
\author[2]{Lei Han}
\author[2]{Gianluca Demartini}
\author[1]{Stefanie Rinderle-Ma}

\affil[1]{TUM School of Computation, Information and Technology, Technical University of Munich, Garching, Germany,
\{catherine.sai, stefanie rinderle-ma\}@tum.de}
\affil[2]{School of Information Technology and Electrical Engineering, University of Queensland, Brisbane, Australia, 
shazia@itee.uq.edu.au, \{l.han, g.demartini\}@uq.edu.au}
\date{}

\begin{document} 
\maketitle
\abstract{Organizations face the challenge of ensuring compliance with an increasing amount of requirements from various regulatory documents. Which requirements are relevant depends on aspects such as the geographic location of the organization, its domain, size, and business processes. Considering these contextual factors, as a first step, relevant documents (e.g., laws, regulations, directives, policies) are
identified, followed by a more detailed analysis of which parts of the identified documents are relevant for which step of a given business process.
Nowadays the identification of regulatory requirements relevant to business processes is mostly done manually by domain and legal experts, posing a tremendous effort on them, especially for a large number of regulatory documents which might frequently change.
Hence, this work examines how legal and domain experts can be assisted in the assessment of relevant requirements. For this, we compare an embedding-based NLP ranking method, a generative AI method using GPT-4, and a crowdsourced method with the purely manual method of creating relevancy labels by experts. The proposed methods are evaluated based on two
case studies: an Australian insurance case created with domain experts
and a global banking use case, adapted from SAP Signavio's workflow
example of an international guideline. A gold standard is created for
both BPMN2.0 processes and matched to real-world textual requirements from multiple regulatory
documents. The evaluation and discussion provide insights into strengths and weaknesses of each method regarding applicability, automation, transparency, and reproducibility and provide guidelines on which method combinations will maximize benefits for given characteristics such as process usage, impact, and dynamics of an application scenario. }

\section{Motivation}

In our globalized world, organizations are faced with keeping track of an increasing amount of requirements from various sources in order to stay compliant. Yet, identifying relevant requirements for a business process from vast amounts of documents requires extensive manual work by highly qualified legal and domain experts \cite{DBLP:conf/re/GordonB14,DBLP:conf/dsit/SchumannMG22}.  
Recent work \cite{SWFR23} defines $13$ Regulatory Compliance Assessment Solution Requirements (RCASR) where RCASR2 -- RCASR8 are investigated in order to provide a fine-granular assessment of compliance between regulatory documents and their realizations, e.g., handbooks. In this work, we investigate RCASR 1 (regulatory document relevance) and RCASR 2 (content relevance) by identifying which regulatory documents a company needs to comply with and the identification of which part of the document is relevant for a company. As businesses become increasingly diverse we further analyze the relevance within the business by not only identifying if a regulatory text is relevant for the entire organization but for a specific process, the sub-processes within the process, and the tasks and throwing events (as these are the active elements) within each sub-process (cf. Fig. \ref{fig:approach_overview}).

Our work aims to develop novel technological solutions to help reduce compliance burdens and breaches. Relevance identification is, even at the process level a challenging and high-stakes task that is currently performed manually by experts \cite{DBLP:conf/re/GordonB14}. 
While it is difficult to make the identification fully automatic,
we study if crowd workers are capable of completing (partial) identification tasks, as they have been proven to be able to create high-quality annotations in complex data problems \cite{estelles2012towards,zuccon2013crowdsourcing,demartini2017introduction}.
Additionally, we evaluate the capabilities of two automated methods: an embedding-based NLP ranking and generative AI as a possible aid for the experts and crowd to better cope with the vast amounts of regulatory texts.
Specifically, the study focuses on the design of a hybrid system to aid domain experts in their compliance assessments. We analyze for which process level granularity (Fig. \ref{fig:approach_overview}) relevance can be identified in sufficient quality and how the selected methods can be used and possibly combined to best support the human experts. 
 
Consequently, this work offers the following contributions:
\begin{itemize}
  \item publicly available data set with two real-world use cases as BPMN2.0 models, process descriptions on all three levels (Fig. \ref{fig:approach_overview}), and the mapped open-access regulatory text passages with their relevancy label for all process levels, aiming to facilitate further research in the area
  \item analysis of the feasibility of state-of-the-art NLP methods to retrieve regulatory relevant texts in a complex legal and business setting for multiple levels of business process detail
  \item analysis of the feasibility of generative AI to answer questions about regulatory relevance in a complex legal and business setting for multiple levels of business process detail
  \item analysis of the feasibility of crowdsourcing to identify relevancy in a complex legal and business setting for multiple levels of business process detail
  \item a novel approach for the identification of relevance for specific process aspects for natural language regulatory requirements, comparing multiple methods and possible combinations of these, resulting in scenario-based application recommendations
\end{itemize}

The paper is structured as follows: The analyzed aspects and approach are presented in Sect. \ref{sec:studydesign}, study design. This is followed by Sect. \ref{sec:methods}, describing the methods used for the approach. Section \ref{sec:implementation} provides the study implementation details. Evaluation and discussion of the findings are presented in Sect. \ref{sec:evaluation} and \ref{sec:discussion}. Section\ref{sec:relwork} discusses related work, followed by Sect. \ref{sec:conclusion} concluding the paper.

\section{Study Design}
\label{sec:studydesign}

\subsection{Analyzed Aspects}
\label{subsec:analyzed_aspects}

The diversity of \textbf{regulatory documents} \cite{DBLP:conf/dsit/SchumannMG22} should be reflected in the data set by containing regulatory text paragraphs from multiple documents of varying origins. Aside from the structure, level of detail, and similar aspects, in terms of relevance identification documents can i.a. vary depending on their geographic region (e.g., country) and domain of applicability. This study includes regulatory documents from different countries and domains as well as domain-independent documents. 
Additionally, we differentiate between external documents (i.e., laws, regulations, directives) that originate from outside the business and internal documents that only apply to the business that created them. All internal documents are automatically business-relevant, but might not be relevant for the process at hand. A further aspect is the availability of the documents. The majority of internal documents is a business secret or at least not publicly available. Due to the crowd-working task and to increase transparency for this publication, only publicly available (meaning open access) documents were included in the analysis. 
Therefore the following 3 groups of regulatory documents are represented in the study, reflecting different combinations of relevance:

\begin{itemize}
    \item \textbf{Group A} contains internal and external documents that are both, business-relevant and 
    \underline{process-relevant}.
    \item \textbf{Group B} contains internal and external documents that are \underline{business-relevant} and process-irrelevant.
    \item \textbf{Group C} contains external documents that are \underline{business-irrelevant} and process-irrelevant.
\end{itemize}

A random selection of all textual content paragraphs from the described documents is extracted, excluding table of contents, tables, and definitions. As the regulatory text passages are extracted from different documents, some context about the regulatory document (title, section title, subsection) is included together with the document meta-information (applicability) mentioned above.

\begin{figure}[ht!] 
    \centering
    \includegraphics[width=0.99\textwidth]{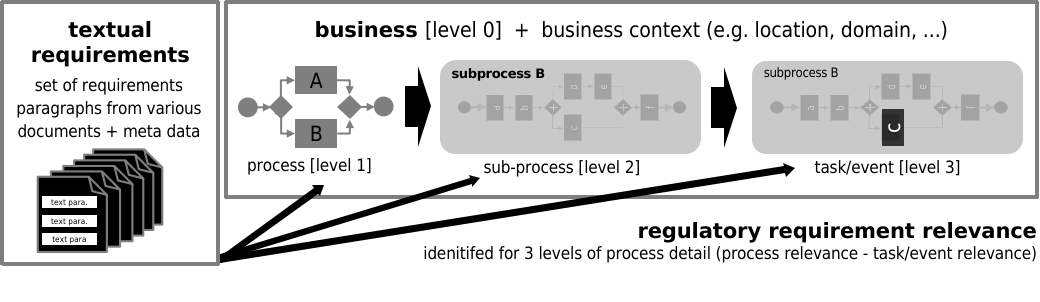}
    \caption{Overview of study aspects analyzed}
    \label{fig:approach_overview}
\end{figure}

The \textbf{relevance judgement} is analyzed for different levels of detail, level 0 being the most general and level 3 the most specific. Level 0 identifies relevance for the business as a whole, while level 1 assesses if a regulatory requirement is relevant for a specific process within the business. Even more detail is provided by level 2, which identifies if a requirement is relevant for a sub-process within a process. Finally, level 3 displays the deepest aspect of a process by identifying relevance for a specific task or throwing event. Through this, we want to analyze if the methods are feasible for all or only certain levels of relevance identification and identify the necessary circumstances for applicability.

Furthermore, interviews with domain experts led us to distinguish two kinds of relevance: 
\textsl{Compliance relevance} is a description of an action an organization has to fulfill in order to be compliant with a regulation (relevance in the stricter sense).
\textsl{Informative relevance} applies, when texts contain information related to the process but do not require a clear action by the organization (e.g. customer obligations in order to fulfill policy requirements).

For the \textbf{business} side, \textbf{processes} with at least one, better multiple relevant regulatory documents should be chosen. The processes are visualized in BPMN2.0 and contain textual descriptions at all three process levels (level 1-3, cf. Fig. \ref{fig:approach_overview})
as \textsl{``textual process descriptions are widely used in organizations''} and hence \textsl{``provide a valuable source for process analysis, such as compliance checking''} \cite{DBLP:journals/is/AaLR18}. 

Four methods are applied for the identification of a relevant text passage from a regulatory document for a given process of an organization. If the regulatory text is relevant for the business process, the relevance of the sub-processes and finally the specific tasks is also identified. This way we assess the relevance identification capabilities for three levels of detail (cf. Fig. \ref{fig:approach_overview}). As this study's focus lies on identifying relevant regulatory texts for processes (and their more detailed aspects), level 0 (business relevance) is not explicitly analyzed. However, if a regulatory text is relevant for a business process (level 1) it is consequently also relevant for the business (level 0). 

The business information includes textual process descriptions for all three process levels (Fig. \ref{fig:approach_overview}), as well as also basic information about the business (location, domain, size). Additionally, for the crowd study a visual BPMN2.0 (\url{bpmn.org)} representation of the root process is included. If the BPMN2.0 model or any textual descriptions are not available, they need to be created manually by domain experts, possibly aided by generative AI.
The presented design ensures that the approach is validated by sources from various origins.

\subsection{Analysis Approach}
\label{subsec:analysis_approach}

\begin{figure}[ht!] 
    \centering
    \includegraphics[width=0.6\textwidth]{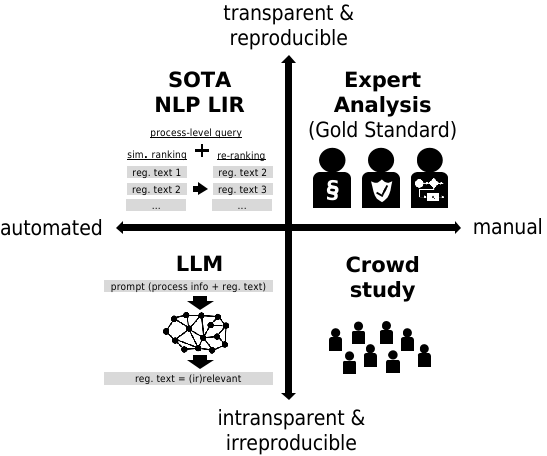}
    \caption{Overview of methods applied for process relevance identification}
    \label{fig:method_overview}
\end{figure}

The study aims at improving the relevance identification of regulatory texts for business processes. This task is currently performed manually by an extensive expert analysis of potentially relevant regulatory texts. One dimension for improvement that the study is capturing is therefore the direction towards a (semi-)automated approach, shown in the horizontal axis of Fig. \ref{fig:method_overview}. Furthermore, the improvement deals with the regulatory compliance of businesses. This is a critical field where violations are quickly resulting in a big impact on the business, often both financially and in reputation aspects. Therefore, the proposed support for the experts also needs to be understandable to a certain degree: how does a method reach its relevance judgment? Additionally, the legal relevance judgment should be consistent in order to be reliable, meaning a method should reach the same relevance judgment when given the same input. This dimension is measured by the vertical axis of transparency and reproducibility (cf. Fig. \ref{fig:method_overview}).

The \textbf{Expert Analysis} is the basis for the approach as on the one hand it is the current standard way to perform the relevance judgment and on the other hand, it is necessary to create the gold standard in order to perform and evaluate the other methods in a quantitative manner. As stated before this method is purely manual and also transparent and reproducible as the experts come to reasoned relevance judgments and document these. 

As a second method, our approach analyses the applicability of state-of-the-art natural language processing legal information retrieval (\textbf{SOTA NLP LIR}), ranking the regulatory texts according to their relevance for the process query. This can be completely automated and still offers full reproducibility and is also explainable and transparent in the way it calculates the relevance ranking based on word embeddings. 

Large language models (\textbf{LLM}s) are currently considered the best supportive system for various tasks that were performed manually thus far and have also shown great results for classifying texts (cf. Sect. \ref{sec:relwork}) this study analyses their capability to support the relevance identification for business processes. Similar to the SOTA NLP LIR method, LLMs only need a well-designed prompt to perform their relevance judgment which can be automatically generated from the process information, they are thus a completely automated solution. However, the LLM relevance judgment can differ if given the same prompt multiple times and it is not clear how the relevance is calculated.

For the fourth method, we analyze if the relevance judgment could be supported in a \textbf{Crowd study} by untrained workers, given a specific task on a complex matter. As experts are rare and expensive resources, crowd studies have shown to be a valuable aid or even alternative for multiple tasks currently performed by skilled workers (cf. Sect. \ref{sec:relwork}). However, it is not clear how the crowd workers reach their relevance judgment and if they would come to the same conclusion if given the same information a second time. Thus, they are considered a manual, intransparent, and irreproducible method.  

Without defined business processes use cases, a collection of potentially relevant regulatory text passages and their relevancy matching it is not possible to test the SOTA NLP LIR, LLM or Crowd Study method. The three newly presented methods depend on the information retained in the Expert Analysis. Thus, the three methods (SOTA NLP LIR, LLM, Crowd study) for the application of relevance judgment will be evaluated based on the gold standard (Expert Analysis) in quantitative terms (cf. Sect. \ref{sec:evaluation}) and the qualitative dimensions of automation and transparency/reproducibility (cf. Sect. \ref{sec:discussion}). 
However, the Expert Analysis is also an independent method to determine relevance by itself, namely the method currently used by the industry, and is thus analyzed as one of four methods in this study.

\section{Methods}
\label{sec:methods}

In the following, the conceptual background of the four methods depicted in Fig. \ref{fig:method_overview} is described in more detail. 

\subsection{Expert Analysis}
\label{subsec:expert_analysis}

\cite{DBLP:conf/re/GordonB14} compare the ability of domain experts, legal experts, and laypersons (non-experts) to assess similarity and interpret legal requirements and definitions. Legal experts achieve the highest rates, followed by domain experts.
\cite{DBLP:conf/dsit/SchumannMG22} conduct interviews with five business auditors in order to assess how the identification of relevant regulatory documents for an internal audit is performed in organizations. According to \cite{DBLP:conf/dsit/SchumannMG22} at the beginning of an audit, it is not clear which and how many regulatory documents need to be considered for compliance. The collection of documents also increases iteratively as one document refers to another document. Thus, the \textsl{``search for [...] regulatory documents is often performed exploratively and can take some time''} \cite{DBLP:conf/dsit/SchumannMG22}. For this study, we interviewed three domain experts from the insurance industry about their actions to ensure regulatory compliance of their processes. The insurance company currently works with an external provider with a big team of legal experts that identifies relevant regulatory text passages per quality unit of a process. Although we did not speak with the legal experts directly, we received their insights implicitly from the domain experts.

The combined challenges \textbf{[C]} elicited from  \cite{DBLP:conf/dsit/SchumannMG22, DBLP:conf/re/GordonB14} and our interviews are: 
\begin{itemize}
  \item high degree of \textbf{manual} work/ lack of automation approaches \textbf{[C1]}
  \item lack of \textbf{context} as the search for relevant documents is often only based on a few keywords which 
  although the context of the terms \textsl{``plays an essential role''} \cite{DBLP:conf/dsit/SchumannMG22} \textbf{[C2]}
  \item need for \textbf{highly skilled workers} to correctly interpret legal requirements, which are costly and hard to obtain \cite{DBLP:conf/re/GordonB14} \textbf{[C3]}
  \item \textbf{decentralized} data (multiple tools, excel sheets, documents) \textbf{[C4]}
  \item \textbf{variety of regulatory documents } which \textsl{``differ greatly in terms of subject matter as well as structure, vocabulary, and level of detail''} \cite{DBLP:conf/dsit/SchumannMG22} \textbf{[C5]}
\end{itemize}

Based on these challenges and the dimensions \textsl{automation} and \textsl{transparency/reproducibility} (cf. Fig. \ref{fig:method_overview}), we select the other methods to be compared in the study. The aim is a semi-automated approach that supports the experts, knowing that full automation is neither technologically feasible nor desired from a responsibility and reliability point of view \textbf{[C1+C3]}. We aim at an approach that can deal with various regulatory documents by not being built on custom dictionaries or rules \textbf{[C4+C5]}.  
The identification input also includes as much context information, as far as feasible by the methods nature \textbf{[C2]}.

Furthermore, together with the experts, suitable business processes and regulatory documents (cf. Sect. \ref{subsec:analyzed_aspects})  need to be identified and the gold standard is created. The procedure is, that for each of the regulatory text paragraphs, the relevancy type (irrelevant, compliance relevant, or informative relevant) is decided for the process (level 1). The processes identified as relevant are then further labeled at process levels 2 and 3 (cf. Fig. \ref{fig:approach_overview}).

\subsection{SOTA NLP LIR}
\label{subsec:rule_based_design}

Legal Information Retrieval (LIR) is concerned with identifying the top k relevant documents or case laws. Missing one document is not severe as this would only mean that not each related case from the past was identified to review it for potentially relevant information. This differs from this study's application: it is extremely important not to miss any relevant regulatory texts as this could lead to compliance issues. However, as both applications aim at identifying relevance of legal text and no method exists yet for the business process application, with this method we test the ability of the in recent publications \cite{DBLP:conf/dsit/SchumannMG22, askari2023injecting} most promising NLP LIR approaches.   

There are two main methods to perform the similarity computation for information retrieval: lexical document retrieval systems, which need an exact match, and semantic document retrieval systems, which are able to identify semantically similar terms \cite{DBLP:journals/ijswis/HliaoutakisVVPM06, DBLP:journals/corr/abs-2010-01195}.
Furthermore, recent, related work \cite{DBLP:conf/dsit/SchumannMG22, askari2023injecting} identifies the combination of a lexical and semantic search with a cross encoder re-ranking as most promising to retrieve potentially relevant text passages based on an input query, in our case the process(-events) description text. We, therefore, implement two state-of-the-art NLP LIR methods, to use the better performing method as a baseline approach for retrieving relevant legal texts for a given process and its elements.

The algorithms described in this method are not novel, yet the application this study applies them to is. Other publications apply these or similar methods to e.g., retrieve supporting case laws based on a query case (cf. COLIEE 2023 \cite{DBLP:conf/icail/GoebelKK0SY23}) or \cite{DBLP:conf/dsit/SchumannMG22} for retrieving relevant German regulatory documents for an audit based on key phrases from auditors. Yet, to the best of our knowledge no approach exists, that analyses the applicability of these methods for the retrieval of relevant regulatory texts based on detailed business process information. Our application scenario also poses new challenges compared to the existing use cases as with the complex business process context, we have more information to consider for the query design than, e.g., a 2--3 word key phrase.  

\begin{figure}[ht!]
\vspace{-2mm}
    \centering
    \includegraphics[scale=0.75]{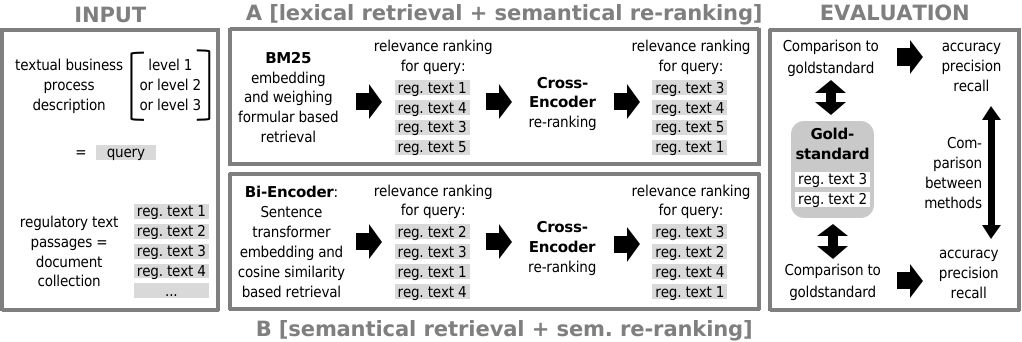}
    \caption{SOTA NLP LIR design inspired by \cite{reimers-2019-sentence-bert, DBLP:conf/dsit/SchumannMG22}}
    \label{fig:sota_nlp_lir_design}
\end{figure} 

Figure \ref{fig:sota_nlp_lir_design} illustrates the steps of the SOTA NLP LIR methods for relevance judgment concerning regulatory business process requirements. 
Method A is based on a lexical search with BM25, re-ranking the retrieved passages with semantic Cross-Encoder. Method B already bases the initial retrieval stage on semantic search with Bi-Encoder and also re-ranks all retrieved passages with the semantic based Cross-Encoder. In the following the elements of the methods are explained briefly. 

\textbf{BM25} ranking function is a lexical document retrieval system. It is based on the ``bag-of-words'' approach in NLP, where the words in a text are considered independent of their position in the text like all the words of a text were put into a bag \cite{DBLP:journals/ftir/RobertsonZ09}. Using measures like term frequency (TF) and inverse document frequency (IDF), the words are transformed in a mathematical representation (also called embedding) and weighed in a certain formula to evaluate their ranking.

\textbf{Sentence Transformer} is a transformer-based embedding framework that, in contrast to BM25, takes the context of the word (e.g. surrounding words) into consideration \cite{reimers-2019-sentence-bert}. Both semantic components (Bi-Encoder and Cross-Encoder) rely on sentence-transformer models. 
The \textbf{Bi-Encoder} \cite{reimers-2019-sentence-bert} passes the texts independently to Sentence-Transformer and compares the resulting embeddings based on cosine similarity. This set-up is less computational intensive (than, e.g., Cross-Encoder) and thus recommended for information retrieval tasks as an initial step to retrieve a pre-selection of potentially relevant texts from a larger collection. 

In comparison, \textbf{Cross-Encoder} \cite{reimers-2019-sentence-bert}, like Bi-Encoder is a semantics based encoder, but \textsl{``Cross-Encoder does not produce a sentence embedding''} \footnote{\url{https://www.sbert.net/examples/applications/cross-encoder/README.html}}. Both texts are passed together to the Sentence-Transformer and the result is a similarity classification between 0 and 1. Cross-Encoders deliver better results in relevancy ranking. However, they are too computationally intensive to use them directly on large document collections.

\subsection{Generative AI Study}
\label{subsec:gpt_design}
For the generative AI Study a zero-shot approach is chosen, meaning only the first response from GPT-4 is considered, no regenerate responses or further information is given. Zero-shot approaches have performed well for multiple tasks \cite{DBLP:conf/nips/KojimaGRMI22} and for the alternative of few-shot learning, use case specific examples would need to be created which would be against our design principle of creating an approach as general and automated as possible. In this kind of setup, the prompt used for the generation needs to be carefully designed.

\begin{figure}[ht!]
\vspace{-2mm}
    \centering
    \includegraphics[scale=0.80]{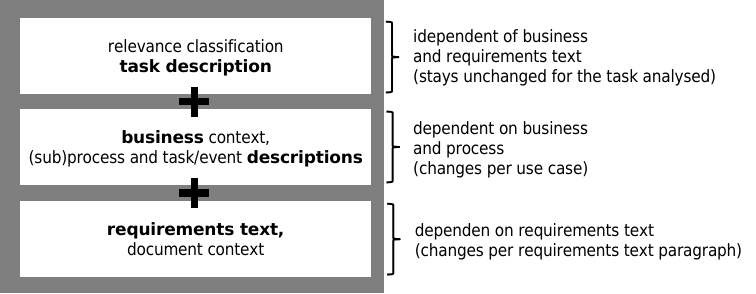}
    \caption{Prompt design components in this study}
    \label{fig:prompt_design}
\end{figure}

\cite{DBLP:journals/corr/abs-2212-02199} propose a ``legal prompt stack'', where the prompt should consist of the information needed to answer the task (in their case the legal document text), in our case the business context and requirements text. Additionally, a description of what the model is expected to do is required. In \cite{DBLP:conf/bpmds/BuschRSL23} prompt engineering for BPM applications is discussed, stating that a task description is needed for zero-shot design and that prompt templates are task specific. They also state that is an open challenge to create prompt templates for business processes and their complex representations. 

Based on these existing prompt proposals and our application setting, we design the prompt consists of 3 parts, as shown in Fig. \ref{fig:prompt_design}: 
1) a description of the task to be performed, 
2) information about the business process: organizational context, the process description, sub-process descriptions and event descriptions, 
3) information about the regulation: one paragraph from a regulatory text and regulatory document context information.
Part 1) stays the same for all runs, 2) is the same per process and 3) is changed for each run with a different regulatory document text. 
The prompt is created in three iterations. After each iteration, the results are evaluated on a small test set of 12 regulatory texts. Depending on this analysis the prompt is adjusted and evaluated again.

\subsection{Crowd study}
\label{sec:design_crowd}

As crowdsourcing has provided large opportunities to generate manual annotations at scale \cite{demartini2017introduction}, we study the possibility of harvesting the ``wisdom of crowds'' in relevance assessment for regulatory
requirement text for business processes.
Hence, we design a crowdsourcing task that allows crowd workers to provide manual annotations. Considering the complexity of matching each piece of regulatory text to low-level business process activities, we design the task in a two-phase fashion:

\begin{itemize}

\item (Phase-1) Each worker is asked to assess the relevance of a given piece of regulatory requirement text with respect to a given business process, and, if relevant, identify which (one or more) sub-processes in the process the given text is most relevant to; and

\item (Phase-2) Each worker is asked to identify which (one or more) particular activities in the sub-process the given text is most relevant to, and the type of relevance, i.e., informative relevance or compliance relevance.

\end{itemize}

\begin{figure}[ht!]
\vspace{-0.2cm}
\centering
\includegraphics[width=0.9\textwidth]{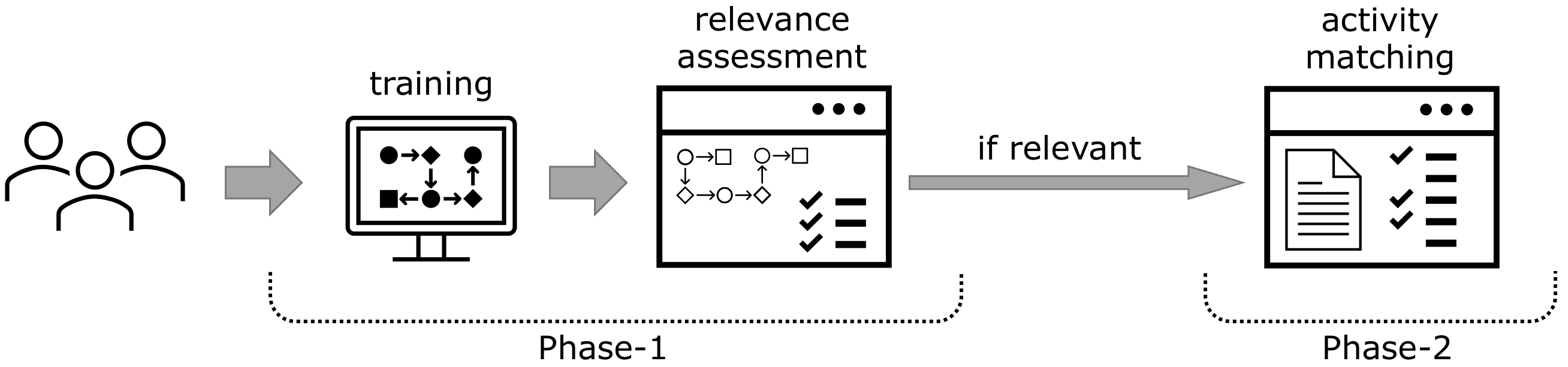}
\caption{Procedure of the crowd study}
\label{fig:crowd_procedure}
\vspace{-0.2cm}
\end{figure}

Figure~\ref{fig:crowd_procedure} shows the overall procedure of our crowdsourcing experiments. As a process model can aid our understanding of the overall business process, we first introduce the BPMN2.0 notations that were used through two simple examples: organizing a meeting, and restaurant handling an order. We provide a textual description for each of these notations. Then, the participating crowd workers are presented with the actual task of the relevance assessment.

In the task, we present both the textual description as well as the root process in BPMN2.0 for the studied use cases.
Note that we publish the two-phase tasks one after another. That is, having collected the annotations from the Phase-1 task, we publish the Phase-2 task by using the text that has already been judged as relevant.
In the Phase-2 task, the workers need to match the given text to relevant activities in the sub-processes. 
In both tasks, all workers are required to provide a textual justification for their selection.

\subsection{Comparative Remarks}
\label{sec:comparative_remarks}

All methods need to identify the same amount of relevant text paragraphs. For the automated approaches SOTA NLP LIR and LLMs, more irrelevant text paragraphs are added to the input data to create a situation closer to the actual business setting. 
The two automated methods share the same input information provided in different forms as LLMs are designed to handle large amounts of context data while the existing LIR methods are based on comparably very short queries.

\section{Study Implementation}
\label{sec:implementation}

The created data sets, prompts for the generative AI and the SOTA NLP implementation are publicly available on Github\footnote{\url{https://anonymous.4open.science/r/regulatory_relevance4process-D73C}}.

\subsection{Expert Analysis} 
Following the described criteria (cf. Sect. \ref{sec:studydesign}) and procedure (cf. Sect. \ref{subsec:expert_analysis}) two suitable business processes and a composition of relevant and irrelevant regulatory documents for these are identified and a gold standard is created. For use case 1 the gold standard is manually created and reviewed with three insurance domain experts, for use case 2 the gold standard is created and reviewed by two business process experts.

\subsubsection{Regulatory requirements}
This study analyzes text paragraphs from 7 Australian regulatory documents (e.g., the ``Fair Work Act 2009'') as well as one business internal document that is publicly available. The later document is anonymized for the study (e.g., company name replaced by a placeholder, company address deleted) and is excluded from the published data set to preserve the anonymity of our industry cooperation partner.  

\subsubsection{Business process}
In 2021, \cite{10.1145/3444689} identified $404$ primary studies for their survey in the field of Natural Language Processing for Requirements Engineering. However, only $7$\% of these were evaluated in an industrial setting, which \textsl{``highlights a general lack of industrial evaluation of NLP4RE research results''} \cite{10.1145/3444689}. 
Therefore, for this study, we cooperate with an industry partner from the insurance domain for use case 1. 

\paragraph{Use case 1: travel insurance claims}
The use case of travel insurance claims needs to comply with multiple regulatory documents. As neither a process model nor textual process descriptions existed at our industry partner, we create them together with the domain experts in an iterative manner.

\begin{figure}[ht!] 
    \centering
    \includegraphics[width=0.99\textwidth]{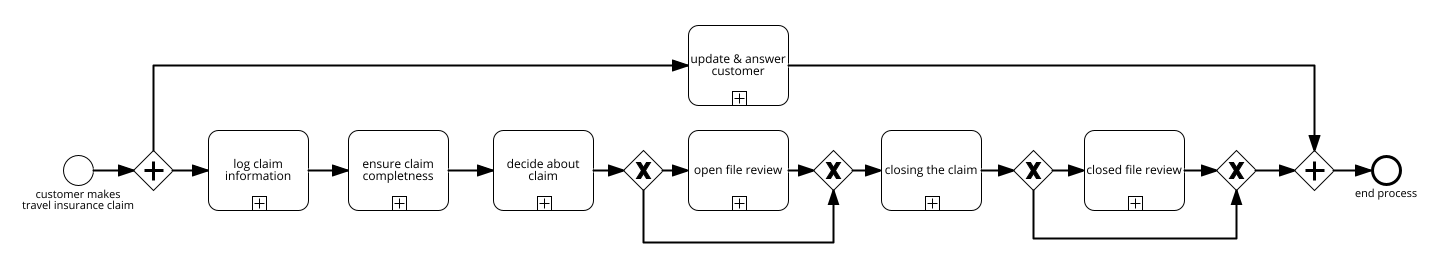}
    \caption{Travel insurance claim process modeled with domain experts from an Australian insurance company, modeled with SAP Signavio}
    \label{fig:use_case_1_claims_root_process} 
\end{figure}

The process deals with the handling of travel claims customers might make against their travel insurance company. It can be grouped in 7 sub-processes including the registration of the claim at the insurer, ensuring all required information is present, communicating with the customer, deciding about the eligibility and calculating a payout amount as well as random quality checks both during and after the payout. Figure \ref{fig:use_case_1_claims_root_process} displays the root process.

\paragraph{Use case 2: know your customer}

\begin{figure}[ht!] 
    \centering
    \includegraphics[width=0.99\textwidth]{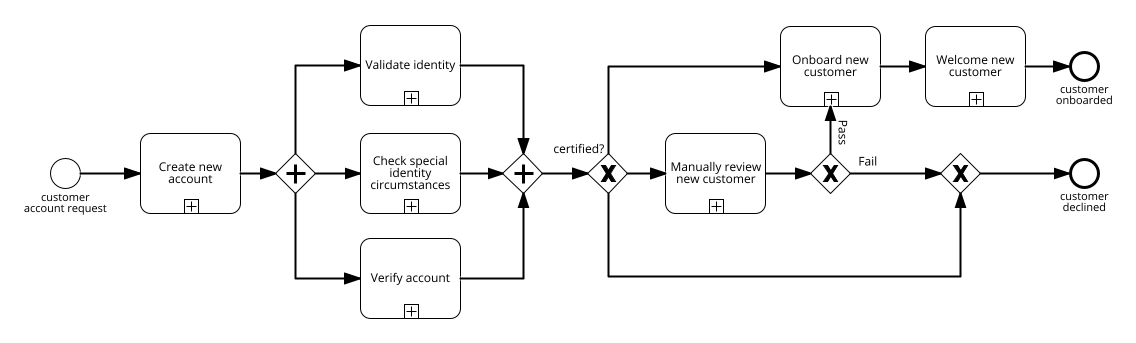}
    \caption{KYC process model based on SAP Signavio workflow example}
    \label{fig:use_case_2_KYC_root_process} 
\end{figure}

The Know Your Customer (KYC) use case of the banking industry is chosen as it is an international standard that needs to comply with different regulatory documents depending on the application country which made it an interesting case for our application. The BPMN2.0 model (cf. Fig. \ref{fig:use_case_2_KYC_root_process})  is based on an SAP Signavio workflow of KYC\footnote{\url{https://www.signavio.com/de/workflow-beispiele/bankingknow-customer-kyc/}}, enriched with textual descriptions created by academic staff and generative AI.

KYC describes a process in the financial domain to ensure a potential customer's identification and due diligence procedures\footnote{\url{https://corpgov.law.harvard.edu/2016/02/07/fincen-know-your-customer-requirements/}}. When a new bank account is created, personal information needs to be reviewed and validated through various checks and confirmations. Depending on the analysis, the new customer might be manually reviewed, following a set of guidelines for acceptability and risk level before being on-boarded and welcomed or declined.

\subsection{SOTA NLP LIR} 
The two embedding-based ranking methods are implemented as described in Sect. \ref{subsec:rule_based_design}. 
Table \ref{tab:input_automated} illustrates the input data used. For the prompt, the business process descriptions were used, e.g. for use case 1, level 3: the 31 task descriptions were used individually as prompt to rank the 489 regulatory text passages (of use case 1) in their relevance for the analyzed task description. 

\begin{table}[htp] 
\footnotesize
\caption{Input data by process level and regulatory text group for the two automated methods}
\centering
\begin{tabular}{lccc|cc|c|c|c}
\toprule
& \multicolumn{3}{c|}{\textbf{Business data}} & \multicolumn{5}{c}{\textbf{Regulatory text}} \\
& Lvl 1 & Lvl 2 & Lvl 3 & \multicolumn{2}{c|}{10\% Grp. A} & 45\%  & 45\% & (100\%) \\
Process & process & sub-pr. & tasks & Compl. rel. & Inform. rel. & Grp. B & Grp. C & total \\\midrule
1: Insurance & 1 & 7 & 31 & 21 & 28 & 220 & 220 & 489 \\
2: Banking & 1 & 7 & 19 & 24 & 7 & 140 & 140 & 311 \\
\bottomrule
\end{tabular}
\label{tab:input_automated} 
\end{table}

\subsection{GPT-4}
The prompts are created based on the application task, as well as the information about the processes and regulatory documents from the use cases as described in Sect. \ref{subsec:gpt_design}. 
After the initial generation, for iteration 2, the task description is enriched with the information to only match clear relations between regulatory text and business process. Additionally, the information about the regulatory text is extended with the automatically generated passages about the regulatory documents content and applicability. 
For iteration 3, the changes from iteration 2 are kept and the task description is extended by the information that recall is the most important measurement for this task.
The final prompts contain between 1500-2300 words, depending on the business case and requirements text.
After three iterations, the resulting prompt is run on all $489$ for case study 1 and $311$ for case study 2 and evaluated against the gold standard. The input data used for the prompts is the same as for the SOTA NLP LIR method and shown in Tab. \ref{tab:input_automated}.

\subsection{Crowd Study}
\label{sec:implement_crowd}

\begin{figure}[ht!]
    \centering
    \includegraphics[width=\textwidth]{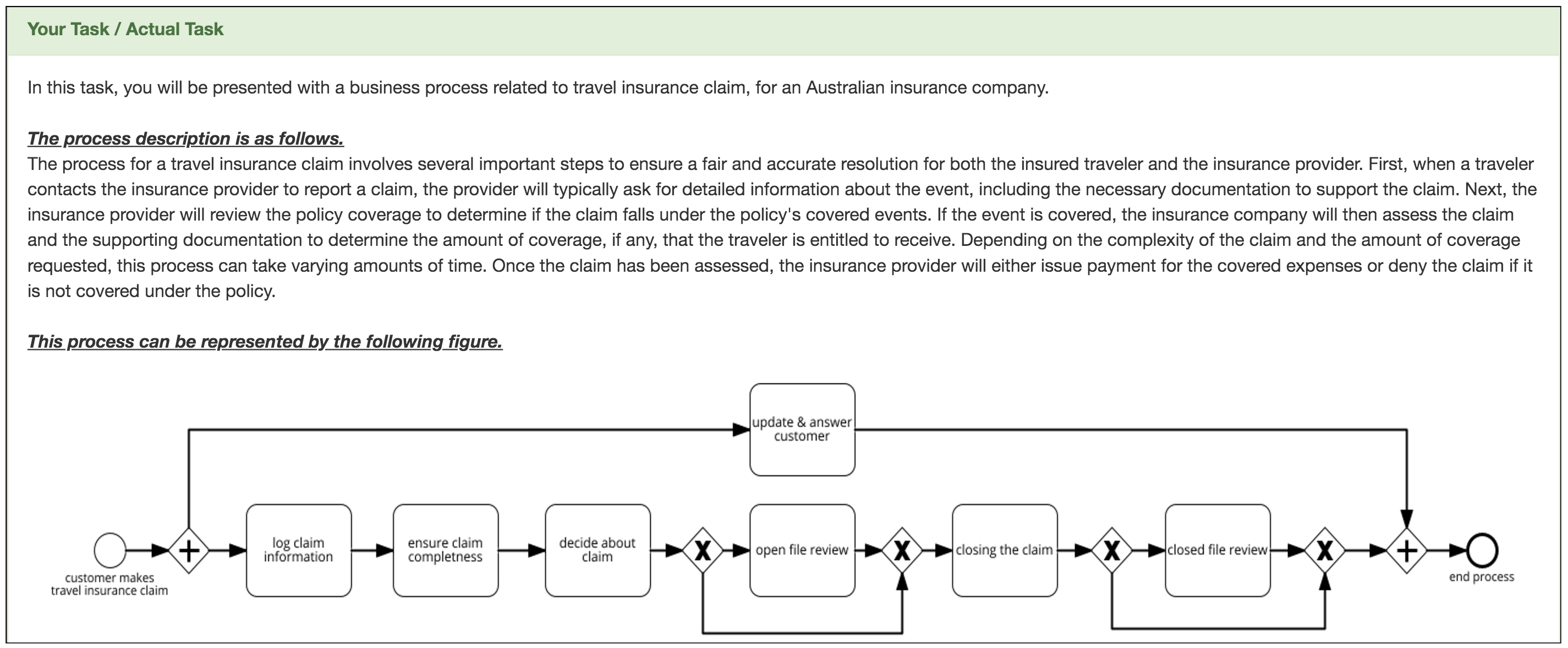}
    \caption{Simplified example of process introduction in crowd study}
    \label{fig:example_crowd_info1}
\end{figure}

\begin{figure}[ht!] 
    \centering
    \includegraphics[width=\textwidth]{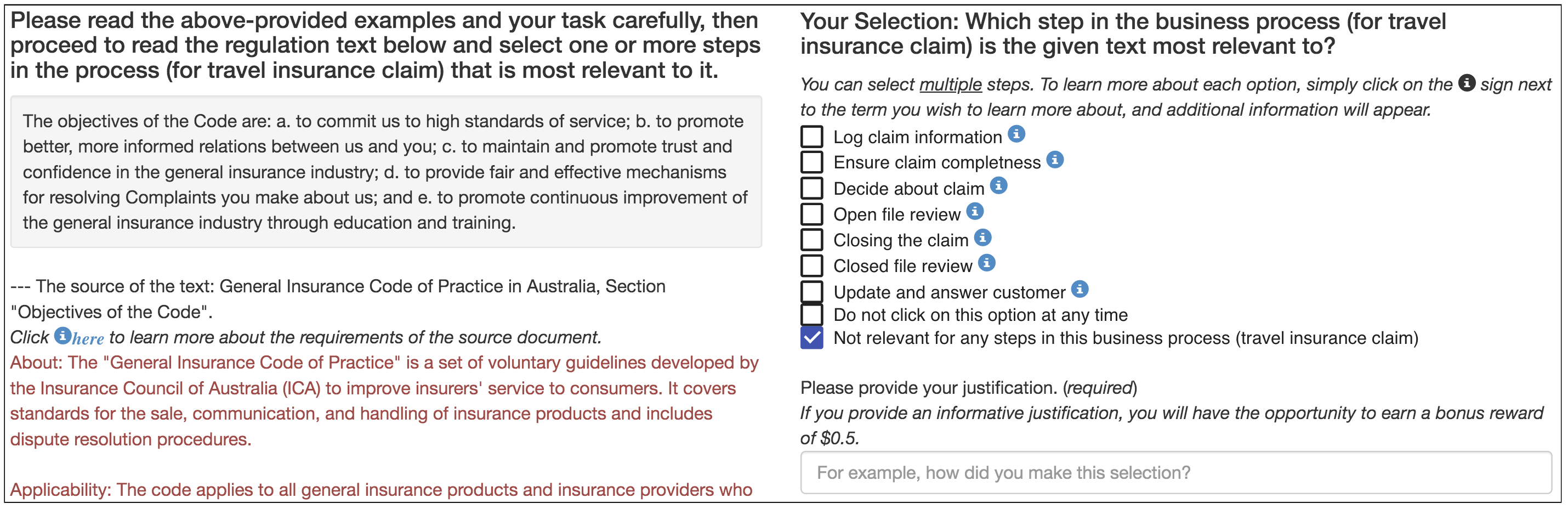}
    \caption{Screenshot of Phase-1 task.}
    \label{fig:example_crowd_2}
\end{figure}

Using the task design described in Sect.~\ref{sec:design_crowd}, we deploy our crowdsourcing experiments on Amazon MTurk (AMT), cf. Fig. \ref{fig:example_crowd_info1} and Fig. \ref{fig:example_crowd_2}. 

We implement three quality checks in both the Phase-1 and Phase-2 task:

\begin{itemize}

\item Two initial test questions to ensure the worker has understood the notations that are used in the examples, which are implemented through the format of binary selection;

\item Attention checks by the embedded JavaScript logger\footnote{All workers have been asked to read and accept our informed consent document before the task where we explain to them about such behavioral action logging. Our study has received Ethics Approval by the review board of the authors' institutions.}, by which among the selection alternatives we embed one option stating ``Do not click on this option at any time'' and we check if the worker has ever clicked on this option; and

\item Semantic dependency of the selected options, where we follow existing crowdsourcing task design that only certain combinations of the answers make sense \cite{kazai2011crowdsourcing} (e.g., it does not make sense if the worker selects both ``Not relevant'' and any labels indicating relevance).

\end{itemize}

These quality checks are performed in a post-experiment manner, which means that all workers are paid at the end of the task and, for those who failed the checks, we re-publish their task on AMT.
To incentivize workers to provide high-quality annotations, we manually check their provided justifications and send them bonus rewards if their justification is informative.
A total of $240$ regulatory paragraphs is analyzed this way by the selected crowd workers, for details about the data composition refer to Tab. \ref{tab:crowd_input_data}. 

\begin{table}[htp] 
\footnotesize
\caption{Crowd study input data by process level and regulatory text group}
\centering
\begin{tabular}{l|ccc|cc|c|c|c}
\toprule
& \multicolumn{3}{c|}{\textbf{Business data}} & \multicolumn{5}{c}{\textbf{Regulatory text}} \\
\midrule
& Lvl 1 & Lvl 2 & Lvl 3 & \multicolumn{2}{c|}{33\% Grp. A} & 33\%  & 33\% & (100\%) \\
Process & process & sub-pr. & tasks & Compl. rel. & Inform. rel. & Grp. B & Grp. C & total \\\midrule
1: Insurance & 1 & 7 & 31 & 21 & 28 & 49 & 49 & 147 \\
2: Banking & 1 & 7 & 19 & 24 & 7 & 31 & 31 & 93 \\
\bottomrule
\end{tabular}
\label{tab:crowd_input_data} 
\end{table}

\section{Evaluation}
\label{sec:evaluation}

The quantitative evaluation of the study is based on accuracy, precision, and recall. Precision and recall are \textsl{``the two most frequent and basic measures for information retrieval effectiveness} \cite{DBLP:journals/ir/Klampanos09}. In contrast to most machine learning evaluations, accuracy is usually not used as a metric for IR systems as normally over $99.9$\% of the documents analyzed are irrelevant to the query, which would lead to very good accuracy by identifying all documents as irrelevant \cite{DBLP:journals/ir/Klampanos09}. However, our application is also similar to a text classification and our input data contains only $90$\% (GPT-4 and SOTA) and $66$\% (Crowd-study) irrelevant samples. Thus, like \cite{DBLP:conf/www/HuangKA23a} in their generative AI study we include accuracy in our metrics.
To calculate these metrics, the values True Positive (TP), False Positive (FP), True Negative (TN), False Negative (FN) are defined as follows:
TP = relevant regulatory text is identified;  
FP = relevant regulatory text is not identified; 
TN = irrelevant regulatory text is not identified; 
FN = irrelevant regulatory text is identified.

Each process level (Fig. \ref{fig:approach_overview}) is evaluated in order to assess the value of each step for each method and discuss possible combinational approaches. 

For the given business application, it is crucial to identify all relevant regulatory requirements. If too many are identified by the automated system, the following step of a human expert-revision can remove these false positives but if relevant texts are not retrieved they cannot be considered in further steps and thus would be left out of process compliance assessments. As \cite{DBLP:journals/ir/Klampanos09} states: \textsl{``various professional searchers such as paralegals [...] are very concerned with trying to get as high recall as possible, and will tolerate fairly low precision results in order to get it.''} Thus, we aim for a recall as close to one as possible, accepting low precision values as a trade-off.

\subsection{Expert Analysis}
Presenting our approach (cf. Sect. \ref{sec:studydesign}) to the domain experts, they confirm that the study results would aid their compliance assessment work. 

For evaluating of the newly proposed methods, we compare their results with the gold standard retrieved by expert analysis. Thus, we assume that the expert analysis did not mislabel anything and has therefore an accuracy, precision, and recall of 1.

\subsection{SOTA NLP LIR}

Table \ref{tab:results_sota_nlp_lir_process_levels} shows the metrics achieved with the two in Sect. \ref{subsec:rule_based_design} introduced methods. To put the following numbers into perspective, the recall of the winning team in the (in Sect. \ref{sec:relwork} mentioned LIR competition) \textsl{COLIEE} 2023 task 1 was 0.41  \footnote{\url{https://sites.ualberta.ca/~rabelo/COLIEE2023/task1_results.html}}.

For use case 1, at process level 1, the BM25 + Cross-Encoder method (cf. Sect. \ref{subsec:rule_based_design}) achieves a recall of 0.43. This is slightly higher than the BI-Encoder + Cross-Encoder method (cf. Sect. \ref{subsec:rule_based_design}), which scores a recall of 0.41. However, for both levels 2 and 3, the BI-Encoder + Cross-Encoder performs clearly better than the BM25 alternative with a recall of 0.39 vs. 0.32 (level 2) and 0.35 vs. 0.23 (level 3). 
As the process relevance levels change from general to more specific (from level 1 to level 3), the recall values for both methods decrease. This might imply that as the relevance levels become more specific, it becomes more challenging for these methods to retrieve relevant information.

\begin{table}[htb!]
\scriptsize
\caption{SOTA NLP LIR results by process level and method}
\begin{tabular}{lc|ccc|ccc}
\toprule
\multirow{2}{*}{process level} & \multirow{2}{*}{method} & \multicolumn{3}{c|}{\textbf{use case 1}} & \multicolumn{3}{c}{\textbf{use case 2}} \\
& & Acc. & Prec. & Rec. & Acc. & Prec. & Rec. \\\midrule
\multirow{2}{*}{level 1: process relevance} & BM25+CE & 0.76 & 0.19 & 0.43 & 0.74 & 0.11 & 0.23 \\
& BI-E.+CE & 0.75 & 0.18 & 0.41 & 0.74 & 0.13 & 0.29 \\ 
\multirow{2}{*}{level 2: sub-process relevance} & BM25+CE & 0.95 & 0.15 & 0.32 & 0.94 & 0.09 & 0.23 \\
& BI-E.+CE & 0.95 & 0.18 & 0.39 & 0.94 & 0.11 & 0.31 \\
\multirow{2}{*}{level 3: task/event relevance} & BM25+CE & 0.97 & 0.09 & 0.23 & 0.97 & 0.08 & 0.14 \\
& BI-E.+CE & 0.97 & 0.13 & 0.35 & 0.97 & 0.13 & 0.24 \\
\bottomrule
\end{tabular}
\label{tab:results_sota_nlp_lir_process_levels}
\end{table}

For use case 2, the BI-Encoder + Cross-Encoder method outperforms the BM25 + Cross-Encoder at every level. At level 1, BI-Encoder + Cross-Encoder achieves a recall of 0.29, while the BM25 + Cross-Encoder method only reaches 0.23. As for use case 1, for both methods, a decreasing trend in recall can be seen as the process level the relevance is identified for becomes more specific. In general, the performance for use case 2 is clearly lower for both methods. This could be an indicator that the process information provided for use case 2 yields less information. 

In both use cases and every process level but use case 1, level 1, the method BI-Encoder + Cross-Encoder outperforms the BM25 + Cross-Encoder method in terms of recall. Also, the results for use case 1, level 1 are very close (cf. Tab. \ref{tab:results_sota_nlp_lir_process_levels}). Therefore, BI-Encoder + Cross-Encoder is the recommended method for the SOTA NLP LIR aiming at business process relevance identification.

\subsection{GPT-4}

For use case 1, GPT-4 identifies all relevant regulatory texts correctly for the process (level 1), resulting in a recall of 1. The lower precision shows that in order to achieve this, multiple false positives have to be accepted. The results by process level (cf. Tab. \ref{tab:results_gpt_4_process_levels}) show that the more detailed the relevance Identification (cf. level 2 and 3), the lower the recall as the differentiation between the aspects of the level becomes less clear. 
For use case 2, one relevant regulatory text is identified as false negative, although by reviewing the text, the reasoning from the generative AI is correct and the label should in fact be irrelevant. The results for both use cases show a very high recall on level 1. Comparing the performance of GPT-4 for the two use cases, a high recall on level 1 can be observed for both use cases. For level 2, GPT-4 performed better on use case 1, while for level 3 it performed better on use case 2.

\begin{table}[htb!]
\caption{GPT-4 results by process level}
\begin{tabular}{lccc|ccc}
\toprule
& \multicolumn{3}{c|}{\textbf{use case 1}} & \multicolumn{3}{c}{\textbf{use case 2}} \\
& Acc. & Prec. & Rec. & Acc. & Prec. & Rec. \\\midrule
level 1: process relevance    & 0.81 & 0.34 & 1.00 & 0.90 & 0.51 & 0.97 \\
level 2: sub-process relevance  & 0.89 & 0.71 & 0.81 & 0.76 & 0.58 & 0.62 \\
level 3: task/event relevance & 0.75 & 0.64 & 0.69 & 0.82 & 0.93 & 0.77 \\
\bottomrule
\end{tabular}
\label{tab:results_gpt_4_process_levels}
\end{table}

\begin{table}[htb!]
\caption{Accuracy by type of regulatory text origin for business process level 1}
\begin{tabular}{lcc}
\toprule
 & \makecell[l]{use case 1} & \makecell[l]{use case 2} \\\midrule
Group A (business relevant, process relevant) & 1.00 & 0.97 \\
Group B (business relevant, process irrelevant) & 0.65 & 0.79 \\
Group C (business irrelevant, process irrelevant) & 0.92 & 1.00\\
\bottomrule
\end{tabular}
\label{tab:gpt_results_by_reg_doc_origin}
\end{table}

For the analysis by type of regulatory text origin (cf. Tab. \ref{tab:gpt_results_by_reg_doc_origin}) an evaluation based on precision and recall would not be insightful, as only group A contains true positive values.
The accuracy for Group B is the lowest, as this is the most ambiguous Group: Group C is quite clearly irrelevant as the whole document is irrelevant to the business; Group A is relevant for business and process and the generative AI is very good at identifying all relevant cases on process level correctly. Group B however is relevant for the business but not for the specific process. Those cases are most difficult to classify as it is partly subjective what is e.g., considered as detailed information relevant to the process or general business requirement irrelevant to the process.  

Concerning the identification of the correct type of relevance (compliance or informative relevance) for the process (level 1), the generative AI identifies $59$\% of the relevant text for use case 1 correctly and $80$\%
of the relevant text
for use case 2.

\subsection{Crowd Study}

We publish the task by the implementation described in Sect.~\ref{sec:implement_crowd}, targeting receiving annotations from 3 different workers who passed our quality checks. Based on the collected data, we perform 3 evaluations:

\begin{itemize}
\item ``unfiltered'', showing the average results from all workers;
\item ``qlt. filter'', by which we consider the first worker submission from those who passed our quality checks (i.e., attention checks and semantic dependency checks, cf. Sect. \ref{sec:implement_crowd});
\item ``qlt. + comb.'', where we adopt the annotations from all the 3 workers who passed our quality checks and considered a piece of reg. text as relevant if any of them assigned a relevance label\footnote{This increases the number of overall positives (true and false) and decreases the number of negatives, which favors the overall objective that we would like to maximize the recall and minimize the omission of relevant text. However, we do not consider ``unfiltered + comb.'' as anyone who failed in quality checks but randomly assigned relevance annotations would eventually make all text be relevant.}.
\end{itemize}

\begin{table}[htb!] 
\footnotesize
\caption{Crowd study results by process level and evaluation options}
\centering
\begin{tabular}{lccc|ccc|ccc|ccc}
\toprule
& \multicolumn{9}{c|}{\textbf{use case 1}} & \multicolumn{3}{c}{\textbf{use case 2}} \\
& \multicolumn{3}{c|}{\textbf{unfiltered}} & \multicolumn{3}{c|}{\textbf{qlt. filter}} & \multicolumn{3}{c|}{\textbf{qlt. + comb.}} & \multicolumn{3}{c}{\textbf{qlt. filter}} \\
& Acc. & Prec. & Rec. & Acc. & Prec. & Rec. & Acc. & Prec. & Rec. & Acc. & Prec. & Rec.\\\midrule
L1   & 0.77 & 0.71 & 0.52 & 0.83 & 0.72 & 0.80  & 0.71 & 0.68 & 0.53 & 0.86 & 0.95 & 0.60\\
L2  & 0.89 & 0.78 & 0.71 & 0.94 & 0.56 & 0.70 & 0.85 & 0.50 & 0.85 & 0.92 & 0.65 & 0.29 \\
L3 & 0.61 & 0.44 & 0.26 & 0.62 & 0.45 & 0.32 & 0.62 & 0.47 & 0.52 & 0.53 & 0.75 & 0.41\\
\bottomrule\\
\multicolumn{12}{l}{L1: Level 1, process relevance; L2: Level 2, sub-process relevance; L3: Level 3, task/event relevance}
\end{tabular}
\label{tab:results_crowd_study} 
\end{table}

The results\footnote{Due to space limit, we show the best performance for Use Case 2.} show that the quality checks are a useful addition to add reliability to crowd study results.
The ``qlt. + comb.'' evaluation delivers best results on level 2 and 3 but performs poorly on process relevance (level 1). 
For use case 1, precision and recall decrease for all options from level 2 to level 3, as the process information becomes more detailed.  
For use case 2, the overall performance of the crowd was lower than in use case 1.
The recall metrics for both use cases and all three levels fail to come close to the desired 100\%, which would mean no relevant text is mislabeled. 

\subsection{Comparative results}

In the following comparison, we consider the results of SOTA NLP LIR achieved with the Bi-Encoder+Cross-Encoder method, as this delivers overall better results than BM25+Cross-Encoder. Similarly, for the crowd study the results received with the ``qlt. filter'' setting are considered in the following. 
Compared to the gold standard created by expert analysis, none of the other three methods achieves recall results over all three process levels and for both use cases that would allow this method to replace the expert analysis but this was already expected and anticipated. GPT-4 performs best and achieves a recall of 1 for level 1 in use case 1 and $0.97$ for use case 2, which is nearly perfect. Both, Crowd study with recall values of $0.8$ and $0.6$, as well as the SOTA NLP LIR with $0.41$ and $0.29$, show worse results. This is similar to sub-process relevance (level 2) and task/event relevance (level 3). For level 2, GPT-4 achieves recall values of $0.81$ and $0.62$, the Crowd achieves $0.7$ and $0.29$, while SOTA NLP LIR only reaches $0.39$ and $0.31$. Finally, for level 3, GPT-4 achieves recall values of $0.69$ and $0.77$, the Crowd achieves $0.32$ and $0.41$, while SOTA NLP LIR only reaches $0.35$ and $0.24$.

For SOTA NLP LIR, GPT-4, and the Crowd, study, the results for use case 1 are overall better than use case 2, and there is a clear tendency that the more detailed the level, the less reliable the relevance judgment. For use case 1, the 3 methods perform best for process level and the results decrease with the level of detail. For use case 2 there are a few exceptions to this as the SOTA NLP LIR method has difficulty with process level 1, while GPT-4 and Crowd workers perform better on level 3 than level 2.  

As described in Sect. \ref{sec:methods} a generative AI like GPT-4 is able to take a lot of context information as input and also has the capabilities to consider these for the relevance judgment. This seems to be advantageous in complex business settings as displayed by the two use cases of this study. 
Crowd workers on the other hand aim to fulfill their task as quickly as possible and thus have limited motivation to consider a large amount of context information. The information for Crowd workers needs to be presented in a well-prepared, condensed manner. The SOTA NLP LIR method was designed similarly to best-performing methods for Legal Information Retrieval tasks. However, those tasks do not have to consider a complex business context for their retrieval. Being based on a simple prompt of the corresponding process level description, the implemented SOTA NLP LIR approach lacks the inclusion of the wider business context, which seems to cause its poorer performance compared to the other approaches. 

Consequently, if the methods are reviewed individually and ranked purely based on their recall results, the expert analysis with its recall of 1 performs best, followed by GPT-4 with recall values between $0.62$--$1.0$, depending on the use case and process level (1-3). The Crowd workers would place 3rd place with recall values between $0.29$--$0.8$, and the SOTA NLP LIR with recall values between $0.24$--$0.41$ would be least favorable.
However, as introduced in Sect. \ref{subsec:analysis_approach}, there are other aspects to be considered for a holistic analysis which will be discussed in Sect.\ref{sec:discussion}.

\section{Discussion}
\label{sec:discussion}

This section discusses the implications of the comparative study regarding applicability, automation, transparency, and reproducibility as well as limitations of the study overall.

\subsection{Implications}

Figure \ref{fig:method_overview} puts the four methods into context of the two dimensions of automation and transparency/ reproducibility which might be key success factors for the application of a particular method. The analyzed methods display extreme cases along the inspected aspects of automation, transparency, and reproducibility (cf. Sect. \ref{subsec:analysis_approach}).
Concerning the \textsl{applicability}, combinations and intermediate stages between the four investigated methods are conceivable. For example, LLM can be custom-trained and business internal. The SOTA NLP approach can also be improved by custom-trained machine-learning approaches for a specific application. If either of these automated methods is combined with the expert analysis, e.g., for an automated pre-selection for the experts, it could decrease the manual workload for reference identification immensely.   

In the following, we evaluate the results from all four methods in comparison and with regard to possible applicability as human-expert aid. Table \ref{tab:method_application_scenarios} suggests recommended method combinations for selected process scenarios based on characteristics describing the process scenario at hand and specified in each column. The recommendations are based on balancing the need for adaptability, regulatory compliance, and expert oversight.
\textbf{Process usage} indicates the frequency or extent to which a business process is utilized, e.g., a core process that is executed every hour or minute or a rarely needed process. Process usage constitutes a measurement of the amount of data available about the process (e.g., for custom training or fine-tuning an AI model) and also gives an indication of the process complexity, as more frequently used processes tend to have more variations and optional tasks. 
The \textbf{process impact} highlights the importance of a high recall in the relevance judgment to prohibit high fines for compliance breaches or loss of good reputation. The actions of a process, for example, could have high-value customer impact or be purely internal with low impact. 
\textbf{Process and regulatory dynamics} reflect how frequently a process or its associated regulations change, while \textbf{regulatory input} measures the number of regulatory documents that need to be considered for the relevance judgment. Both characteristics indicate the need for automation as the amount and frequency of manual work would be high otherwise. 

\begin{table}[htp!]
\footnotesize
\caption{Selected process scenarios with recommended method combinations}
\begin{tabular}{p{1.8cm}|p{2cm}|p{2.2cm}|p{2.2cm}|p{3cm}}
\toprule
\centering \textbf{process usage} & \centering \textbf{process impact} & \centering \textbf{process and regulatory dynamics} & \centering \textbf{regulatory input} & \textbf{recommended method combination} \\\hline

\centering low-high & \centering high &\centering low & \centering low & expert analysis \\
\centering high & \centering high &\centering high & \centering high & SOTA NLP LIR + exp. analysis \\
\centering high & \centering low &\centering high & \centering high & GPT-4 + exp. analysis \\
\centering low-high & \centering low &\centering low & \centering low & Crowd + exp. analysis \\
\bottomrule
\end{tabular}
\label{tab:method_application_scenarios}
\end{table}

In case of a business process with a high impact that has a low change rate both on the process and regulatory side and a low number of regulatory documents to consider, the best approach is the \textbf{expert analysis}. The high impact of the process demands a solution with high transparency and reproducibility. At the same time, the low dynamics and regulatory input allow for a purely manual solution.

If the impact of the process is high and, at the same time, the dynamics and regulatory input are also too high to be handled purely manually, \textbf{SOTA NLP LIR + expert analysis} is recommended. This semi-automated combination delivers the transparency needed for a high-impact process while supporting the manual work by experts with an automated pre-selection. As the SOTA NLP LIR results based on standard encoders are not satisfactory, our recommendation for industry application would be to custom train or fine-tune models to further refine the SOTA NLP LIR method for a specific business application. 

\textbf{GPT-4 + expert analysis} is the method of choice if the process is frequently used, a high number of regulatory documents needs to be considered and these, as well as the process, are highly dynamic. When at the same time the process impact is low, this allows for less transparent automation support. 
Across the three levels and both use cases, GPT-4 delivers the highest recalls and overall best results for relevance identification, compared to the gold standard (expert analysis). The generative AI shows reliable results on the process (level 1) and can thus be a real aid to experts for relevance judgments. GPT-4 can be applied to pre-select the vast amount of regulatory text in the described scenario. The human experts can then review the results identified as relevant in order to remove false positives. For process level 2-3 
the relevance identifications of GPT-4 are not to be trusted by themselves as false negatives occur. Overall, the results demonstrate the potential to be used as insightful reasoning to aid the experts in their final relevance judgment. Further improvements to the results could be achieved by fine-tuning an LLM to a given business and process setting. 

The \textbf{Crowd + expert analysis} method is only recommended in a special scenario. As it is unknown how crowd workers come to their relevance judgments, their results are, similar to the output from GPT-4, not transparent. Hence, crowd workers should only be consulted in low-impact processes, but their results are anyway outperformed by GPT-4. Additionally, due to the nature of a crowd study (cf. Sect. \ref{sec:design_crowd}) as well as the length and complexity of the regulatory and business process information involved in making the relevance judgment, the crowd workers seem not suitable for large amounts of regulatory input or a dynamic setting. 
Therefore, we recommend utilizing crowd studies for indications which process(-levels) are not clearly defined. For example, when 0 out of 3 crowd workers identify a process(-step) as relevant, although it actually is relevant, this indicates unclear process elements and is a valuable insight for the process modeling. Consequently, business processes need to be well defined and documented in order to perform reliable relevance identifications on them. Crowd workers could be a valuable aid for experts in identifying ill-defined processes and improve these, especially for scenarios with low regulatory input and change.  

\subsection{Limitations}
In the following, we discuss selected limitations of the presented study.

\vspace{1mm}
\noindent\textbf{Ambiguity between labels ``compliance -- informative'' and ``informative -- irrelevant'' and general applicability of some regulatory texts:}
False positive regulatory texts are often identified as informative relevant, mainly for texts from group B which are relevant for the business in general, but either considered to belong to another process or as too broad to be related to the specific process. 
Some of these cases could be managed, e.g., by a rule that excludes all regulatory texts from being process-relevant if GPT-4 identifies them as relevant for 3 or more sub-processes. 
The differentiation between what is business or process relevant is challenging, as well, even for academic staff without knowing the entire business process landscape or consulting with domain experts. 

\vspace{1mm}
\noindent\textbf{Distinguishing the relevance for a given process from other, related processes in the business:} For use case 1, the domain experts shared, that there are separate processes with detailed procedures concerning e.g. ``quality assurance'' or ``handling of complaints''. However, as quality and complaints are also mentioned tasks of use case 1, the 2 methods falsely label texts as relevant for those tasks. 
All text from the Privacy Principle regulatory document, for example, are labeled as relevant by GPT-4 for the use case 1 process. This is understandable as privacy principles certainly are important in every process concerned with customer or third-party data, but it is not the concern of the travel insurance claims process to ensure privacy principles. It is relevant for the business (level 0) and for a data security and privacy process the business hopefully has in place but it is not relevant for the process use case 1 as defined by the business domain experts. 

\vspace{1mm}
\noindent\textbf{Standalone ability of extracted regulatory texts:}
Legal texts often include references to other sections and paragraphs within the same document and across documents. As our study is designed to evaluate the relevance of each text paragraph, independently of each other, information exploiting references is currently not accessed, neither by GPT-4 nor the crowd workers although it might provide additional aid in the relevance assessment. 

\vspace{1mm}
\noindent\textbf{Missing context consideration:}
In most cases, the studied methods consider the context of regulatory text and business process. However, there are cases where the provided metadata gives a clear indication of the relevance but is not included in the judgment.

\vspace{1mm}
\noindent\textbf{Data Privacy issues for closed-access documents:}
Both GPT-4 and crowdsourcing pose challenges in handling secret (e.g. business internal requirements) or payable (e.g. ISO-Norm) documents. For the generative AI, it is possible to opt for excluding the submitted data for training the model. Additionally, the GPT-4 provider plans to offer a Business subscription in the future.  

\vspace{1mm}
\noindent\textbf{Cost:}
No closer consideration of the cost for the 2 methods in relation to the time saved for experts solving the task without aid was performed. \\

\noindent Two further aspects are selected \textit{limitations of the general problem space}.

\noindent\textbf{Subjectivity and level of detail in business process models:} 
Typically, different modelers will create different variants of a model. We made the same experience when working with the insurance domain experts to create the model for use case 1 which took several iterations. 
Moreover, a process model, in particular, at a conceptual level, provides an abstraction of reality. We already worked with three levels of detail, but still details on, for example, the task design are not covered, cf. \cite{DBLP:journals/iam/LooyBPS13} for a more detailed analysis of this challenge.
This leads to uncertainty about whether a regulatory text is relevant or not as it might not be relevant depending on the abstraction levels of text and process model. 

\vspace{2mm}
\noindent\textbf{Comparability between process and regulatory information:}
The challenge of ontological alignment between the business and regulatory texts has been observed in previous studies \cite{DBLP:conf/bpm/SadiqGN07}. Examples include differences in active vs. passive style, level of detail, or length of sentences. Challenging alignments might be also caused by different perspectives, e.g. a regulations being written from the customer perspective, while the process is phrased from the businesses perspective.

\section{Related Work}
\label{sec:relwork}

"[\textbf{Business process}] \textbf{compliance}  is a relationship between two sets of specifications: the specifications for executing a business process and the specifications regulating a business." \cite{DBLP:books/igi/09/GovernatoriS09,DBLP:conf/bpm/SadiqGN07,DBLP:books/igi/09/GovernatoriS09} describe that business processes and business obligations (including laws, regulations) are designed and handled independently of each other and that business process compliance is a complex problem, due to the "scale and diversity of compliance requirements and additionally the fact that these requirements may frequently change" \cite{DBLP:conf/bpm/SadiqGN07}. 
Additionally, the majority of existing business process compliance approaches are based on formalized constraints \cite{DBLP:conf/bpm/SadiqGN07, DBLP:conf/bpm/GovernatoriHSW08, DBLP:journals/isf/HashmiGW16}.

 \cite{DBLP:conf/aicol/DragoniVRG17, DBLP:conf/re/SapkotaAYDB12, DBLP:conf/otm/WinterR18, DBLP:conf/caise/WinterR19}  
extract machine-readable compliance requirements from legal natural language text, \cite{DBLP:journals/sosym/LeopoldAPRMR19} compares process descriptions with model-based process descriptions, \cite{DBLP:journals/is/AaLR18} is concerned with the interpretation improvement of textual process descriptions, and \cite{DBLP:conf/er/WinterARW20} assess compliance of process models with regulatory documents. \cite{DBLP:journals/is/LyMMRA15} defines requirements for business process compliance monitoring. Their approach starts with Compliance Requirements (in form of e.g. laws or regulations) which first need to be interpreted into compliance objectives and then specified into compliance rules. How the Compliance Requirements, meaning the relevant laws, regulations, etc. are identified is not covered.  Thus, all of the above mentioned approaches take relevance as an implicit assumption. 

While \textbf{legal information retrieval (LIR)} identifies relevant data based on a large corpus of legal data, we study relevance based on business process data. 
Information retrieval is based on similarity computation, using e.g., BERT-Variations \cite{DBLP:conf/dsit/SchumannMG22} between the search term and the text corpus \cite{DBLP:journals/ir/Klampanos09}. A search term usually consists of a few keywords, rather than a long, complex prompt as in our case. 
Current research \cite{DBLP:conf/jsai/NigamGB22, DBLP:journals/rss/KimROG22, DBLP:conf/dsit/SchumannMG22,DBLP:conf/jsai/RabeloKGYKS20} in the field of regulatory document ranking and retrieval show, that a \textsl{``combination of lexical and semantic retrieval models leads to the best results''} \cite{https://doi.org/10.48550/arxiv.2108.03937}. 
Major research in this area originates form the Competition on Legal Information Extraction/Entailment (COLIEE) \cite{DBLP:conf/jsai/KimRGYKS22}. 
The tasks about relevant text retrieval are related to our problem. However, the fact, that there are different tasks for case and statute law already indicates how different the approaches are even when retrieving relevant law texts based on other relevant law texts of the same type. 
Another very recent work is concerned with classifying regulatory documents as business-relevant or business-irrelevant \cite{DBLP:journals/nca/DimliogluWBCOBOW23}. Our approach focuses on the identification of the relevant processes within the business and even more detailed, which process parts are affected. 
In summary, no LIR analysis exists to identify relevant requirements for a specific business process or even more specific task of a business process as presented in this work. 

Our application area of \textbf{generative AI} falls into the field of text classification, as we want to classify a regulatory text as relevant or irrelevant, given a process scenario and context information. The literature reviews on large language models and their capabilities conducted by \cite{DBLP:journals/corr/abs-2304-01852} come to the conclusion that \textsl{``ChatGPT has tremendous potential in text classification tasks''}. \cite{DBLP:conf/www/HuangKA23a} studies the classification capabilities of ChatGPT with crowd workers for identifying implicit hate speech in tweets. 
They conclude that ChatGPT can aid the understanding of the experts and provides \textsl{``great potential of ChatGPT as a data annotation tool''} \cite{DBLP:conf/www/HuangKA23a}. As of October 2023, the latest GPT (Generative Pre-trained Transformer) version is GPT-4. OpenAI introduces their model with the promise of \textsl{``human-level performance on various professional and academic benchmarks''} \cite{DBLP:journals/corr/abs-2303-08774}, including applicability to the legal domain through e.g. \textsl{``passing a simulated bar exam''} \cite{DBLP:journals/corr/abs-2303-08774}.
Therefore, as applicability to classification tasks and the legal domain has been shown, but no combination of legal data classification, depending on complex process and context data has been performed yet, we choose GPT-4 as one method for this study.

Over the years, \textbf{crowdsourcing} has provided great opportunities to create large-scale manually labeled ground-truth data to harness the ``wisdom of crowds'' for various applications (e.g., training machine learning models) \cite{kittur2013future,demartini2017introduction}.
On paid crowdsourcing platforms such as Amazon MTurk (AMT), crowd workers usually complete tasks anonymously, and, hence, collecting high-quality data becomes a major challenge. Existing research has explored different ways to guarantee the quality of the collected data, such as answer aggregation models \cite{venanzi2014community,hung2018computing}, suitable workers selection \cite{bozzon2013choosing,difallah2013pick,jagabathula2017identifying}, embedding purpose-designed questions (e.g., gold standard) \cite{oleson2011programmatic,quinn2011human}, and encouraging workers to re-think about the questions \cite{mcdonnell2016why}.
Following this line of research, in our crowd study, we adopt three quality checks in the deployed task, including: (1)~test questions to check if the worker has understood the task, (2)~attention checks to filter those who provide random answers, and (3)~sophisticated task design where only certain combinations of the answers make sense according to \cite{kazai2011crowdsourcing} (implementation details are described in Sect.~\ref{sec:implement_crowd}).
On the other hand, \cite{DBLP:conf/www/DemartiniDC12} and \cite{DBLP:journals/tweb/WilsonSLSSZRSLS19} show that a combination of automated
methods with input from crowd workers can improve the quality of the generated annotations.
Therefore, we also consider combining algorithmic approaches with crowdsourcing to
design our process for annotating a large amount of regulatory documents, for which the first step is to understand, to what extent, such a complex task could be completed by AI, experts and crowds.

\section{Conclusion}
\label{sec:conclusion}

This paper studies how legal and domain experts can be supported by generative AI, embedding-based ranking, and crowdsourcing approaches when assessing the relevance of regulatory documents for business processes, i.e., their business context, their process context, and their tasks. This assessment is challenging due to the volume and variety of regulatory documents. In order to tackle these challenges, the study provides recommendations of method combinations for selected process scenarios with different characteristics, i.e., usage, impact, dynamics, and regulatory input, allowing for and enabling different levels of automation and transparency/reproducibility. Scenarios with high usage, impact, dynamics, and regulatory input, for example, demand for automation to handle the volume of input data and high transparency at the same time. Here a combination of embedding-based ranking and expert analysis is advisable. 
Generative AI can be a great human-aid in relevancy identification, especially on process level and also give indications and consideration ideas for the relevance judgment of the more detailed aspects of a process. However, due to its limitations in terms of transparency and reproducibility, we only recommend its usage in certain scenarios. Under certain circumstances, e.g., if the focus is more on the transparency of the support method or the goal is more towards improving the process documentation for the automated methods, embedding-based ranking and crowdsourcing can also be valuable assets to the relevance judgment for business processes. 
Such a semi-automated approach relieves the burden on experts and overall contributes to improving business process explainability, traceability, and the compliance of processes with textual requirements.

The presented study can be extended in several ways. With respect to pre-processing,
the presented approach does not include the search and scraping for relevant regulatory documents for a given business process, e.g., on the web or in closed-access ISO-Norms. This extension would allow for a holistic automated pre-selection of all regulatory requirements for a process that then needs to be evaluated for relevancy.
Regarding the processing of the documents, 
in future work, one could further investigate how concepts from Legal Information Retrieval could be adjusted to deal with the given business application and compare the results of those lexical and semantical systems with the methods investigated in this study. 
As generative AI methods seem very sensitive to the prompt design and phrasing \cite{DBLP:journals/tacl/JiangXAN20}, multiple prompt runs with, e.g., para-phrases could be considered in the future. 
Finally, the inclusion of a confidence score in the responses (of both methods) could further aid the usability of the results during post-processing. 
The identified matches between regulatory texts and business process could be integrated and visualized within BPMN models, as prior studies \cite{WANG2022101901} that analyzed stakeholders preferred representation of rules referring to business processes, found that rules linked to BPMN models show the best results.

  \bibliographystyle{plain}

\end{document}